\pdfoutput=1
\documentclass[11pt, a4paper, logo, copyright, nonumbering]{roblox}

\makeatletter
\def\@BTrule[#1]{%
  \ifx\longtable\undefined
    \let\@BTswitch\@BTnormal
  \else\ifx\hline\LT@hline
    \nobreak
    \let\@BTswitch\@BLTrule
  \else
     \let\@BTswitch\@BTnormal
  \fi\fi
  \global\@thisrulewidth=#1\relax
  \ifnum\@thisruleclass=\tw@\vskip\@aboverulesep\else
  \ifnum\@lastruleclass=\z@\vskip\@aboverulesep\else
  \ifnum\@lastruleclass=\@ne\vskip\doublerulesep\fi\fi\fi
  \@BTswitch}
\makeatother

\addto\extrasenglish{
}

 {\begin{list}{}%
         {\setlength{\leftmargin}{#1}}%
         \item[]%
 }
 {\end{list}}
 
\reportnumber{001} %

\renewcommand{\today}{}

\title{\centering Cube: A Roblox View of 3D Intelligence}

\author{Foundation AI team, Roblox\footnote{See Appendix for the full author list.}}

\usepackage[authoryear, sort&compress, round]{natbib}
\usepackage{dblfloatfix}
\usepackage{ulem}
\usepackage{caption}
\usepackage{dramatist}
\usepackage{xspace}
\usepackage{pifont} %
\usepackage{multirow}
\usepackage{tcolorbox}
\usepackage{xltabular}
\usepackage{longtable}
\usepackage{hyperref}
\usepackage{amsfonts}
\usepackage{amsmath}
\usepackage{amssymb}
\usepackage{lineno}
\usepackage{multirow}
\usepackage{adjustbox}
\usepackage{graphicx}
\usepackage{wrapfig}

\usepackage[bottom]{footmisc}

\usepackage{CJKutf8}
\usepackage{subcaption}
\usepackage{setspace}

\usepackage{dsfont}
\usepackage{array} %
\usepackage{tabularx} %
\usepackage{xcolor} %
\usepackage{tabularx}
\usepackage{booktabs}

\usepackage{lipsum}  %
\usepackage{multicol} %
\usepackage{tasks}

\definecolor{MyPurple}{RGB}{111,0,255}

\begin{document}

\begin{abstract}
\vspace{-1em}
 {\small 
Foundation models trained on vast amounts of data have demonstrated remarkable reasoning and generation capabilities in the domains of text, images, audio and video. 
Our goal at Roblox is to build such a foundation model for 3D intelligence, a model that can support developers in producing all aspects of a Roblox experience, from generating 3D objects and scenes to rigging characters for animation to producing programmatic scripts describing object behaviors. 
We discuss three key design requirements for such a 3D foundation model and then present our first step towards building such a model.
We expect that 3D geometric shapes will be a core data type and describe our solution for 3D shape tokenizer. We show how our tokenization scheme can be used in applications for text-to-shape generation, shape-to-text generation and text-to-scene generation. 
We demonstrate how these applications can collaborate with existing large language models (LLMs) to perform scene analysis and reasoning.
We conclude with a discussion outlining our path to building a fully unified foundation model for 3D intelligence. Our code and model weights can be found at: \url{https://github.com/Roblox/cube}\footnotemark.
}
\end{abstract}
\maketitle

\footnotetext{We have updated the code and model since the release in March 2025. Please see Section~\ref{sec:v2_updates} for more details.}
\begin{figure}[h!]
\includegraphics[width=\linewidth]{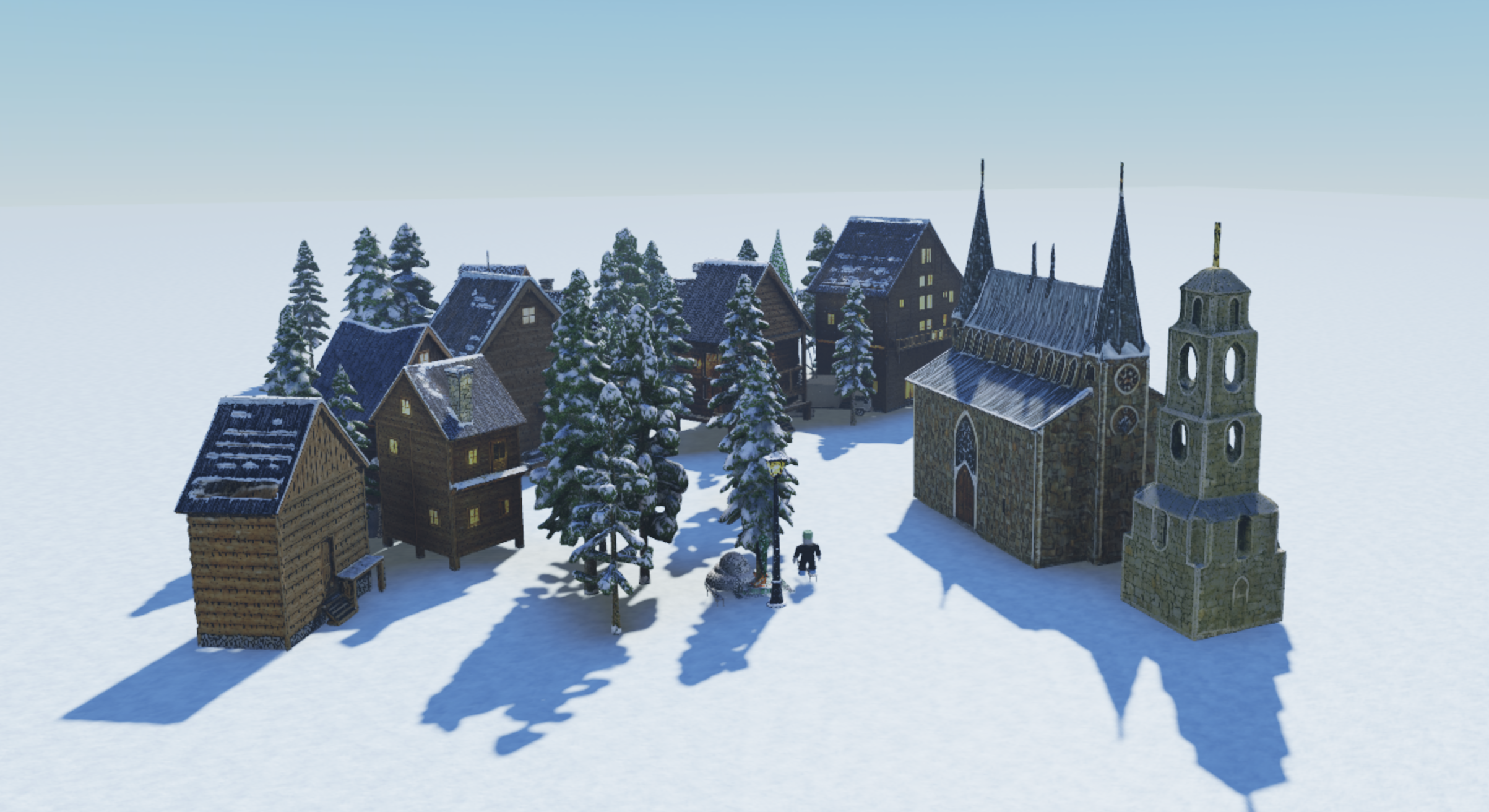}
\caption{{\bf Scene Generation.} We are developing a foundation model for 3D intelligence that will support applications like scene generation. This winter village scene was generated through a multi-turn conversation with our prototype scene generation tool.}
\label{fig:teaser}
\end{figure}

\section{Introduction}

\begin{figure}
    \centering
    \includegraphics[width=\linewidth]{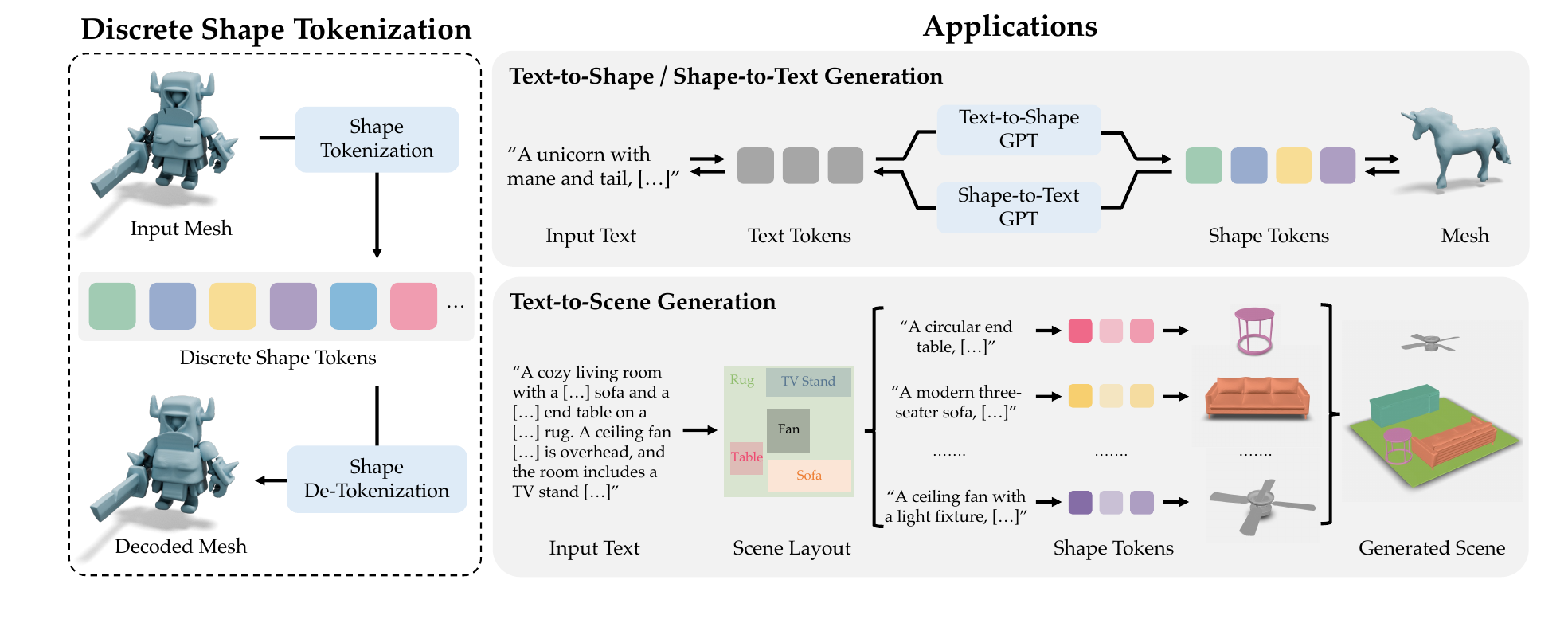}
    \caption{\textbf{Overview.} \small{We present an important step towards the foundation model for 3D intelligence. Specifically, our report focuses on 3D shape tokenization—a technique for converting between shapes and discrete tokens. We also demonstrate how our tokenization scheme enables multiple applications including text-to-shape generation, shape-to-text generation, and text-to-scene generation. } }
    \label{fig:overview}
\end{figure}

Modern generative AI promises to give every human a collaborative assistant for creating visual content. An assistant that can convert simple inputs, such as text prompts, into production-quality visual experiences. 
Today, these assistants typically rely on an underlying foundation AI model that is trained on vast amounts of broad, multi-modal data, and then fine-tuned to perform specific downstream assistive tasks\,\citep{bommasani2022opportunitiesrisksfoundationmodels}. 
Recent foundation models that bridge the domains of text, image, audio and video have demonstrated remarkable reasoning and generation capabilities\,\citep{achiam2023gpt,team2024gemini}.%

At Roblox, our goal is to build a foundation model for 3D intelligence.
We envision this model as the base for a variety of collaborative assistants that will aid developers in producing all aspects of Roblox experiences from creating individual 3D objects (e.g., make a motorbike with wings), to full 3D scene layout (e.g., create a futuristic cloud city), to rigged character avatars (e.g., generate an alien ninja capable of doing wall jumps) to scripts describing object behaviors, interactions and game logic (e.g., make the door open when a player is near it and carrying a gold key). 
We believe that building such a foundation model for 3D reasoning and generation imposes three core design requirements.

\begin{itemize}
\item {\bf Learn jointly from sparse, multi-modal data.} The amount of readily available 3D experience data suitable for training is much less than the data available for training text, image or video models. But 3D data is multi-modal (e.g., geometric meshes, CSG parts, layouts, textures, rigging, programmed scripts) and the modalities are strongly correlated with each other (e.g., an object's texture and placement in a scene often strongly correlate with its shape). %
Thus, our model should jointly learn from all the modalities available in the training data.

\vspace{0.5em}
\item {\bf Handle unbounded input/output size via an autoregressive model.} 3D experiences can be of vastly different sizes, such as a micro-scene containing a few static objects (e.g., a room with a dinner table and chairs) to a multi-apartment building with interactive doors and elevators, to a large-scale city with interacting vehicles and non-player characters. To handle inputs and outputs at each of these scales, our model should be autoregressive with large context length.

\vspace{0.5em}
\item {\bf Collaborate with humans and other AI systems via multi-modal inputs/outputs.}
Humans are familiar with data modalities such as natural language text, images, sketches and video and should be able to express design intents using any of them.
Moreover, existing large language models (LLMs) such as GPT-4o~\citep{achiam2023gpt}, which are trained with vast amounts of multi-modal data including text, programs, images, vector graphics, scene graphs, video, speech, audio, etc. can provide common sense knowledge (i.e., general priors) on a variety of topics.  
Thus, for our model to seamlessly collaborate with humans as well as other LLMs, multi-modal data should be first-class input and output data types.

\end{itemize}

In this report we present our first step towards this foundational model for 3D intelligence. Specifically, we focus on discrete tokenization of the 3D shape, recognizing that geometry will be a central data type of our foundation model. 
We show how our tokenization scheme can be used to build several applications including text-to-shape generation, shape-to-text generation, and text-to-scene generation (Figures\,\ref{fig:teaser} and\,\ref{fig:overview}). We also demonstrate how these applications can collaborate with a text-based LLM to perform scene analysis and reasoning tasks. 
Our code and inference model weights are available at: \url{https://github.com/Roblox/cube}.

\section{Shape Tokenization}
\label{sec:method:vq-vae}

\begin{figure}
    \centering
    \includegraphics[width=\linewidth]{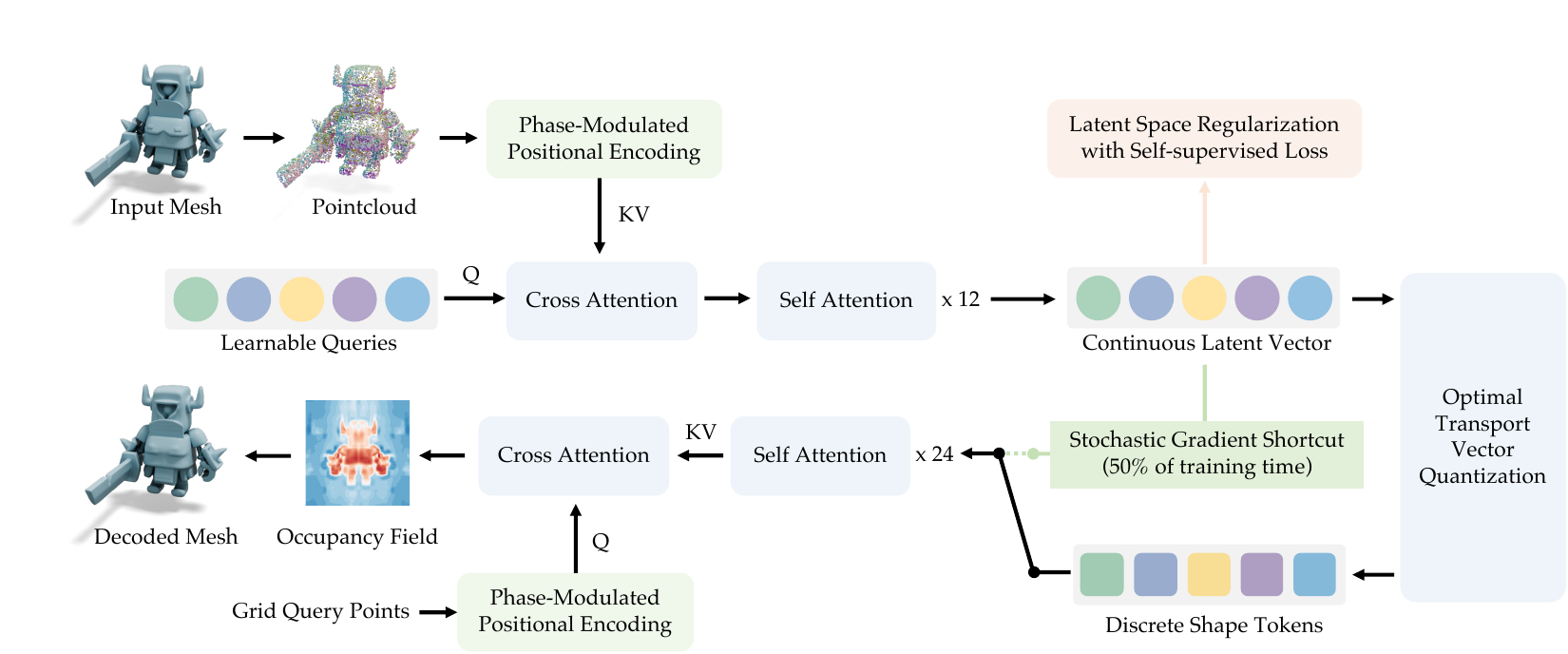}
    \caption{\textbf{Shape Tokenization.} \small{Our pipeline encodes an input mesh into discrete tokens through several steps: (1) We sample points from the mesh surface and embed them using our Phased-Modulated Positional Encoding; (2) A Perceiver-based transformer~\citep{jaegle2021perceiver} encodes these points into continuous latent vectors, regularized with a self-supervised loss; (3) We apply optimal-transport vector quantization~\citep{zhang2024optvq} to convert these vectors into discrete shape tokens;
    (4) These tokens can later be decoded into an occupancy field for mesh extraction. To improve training stability and the reconstruction quality, we also introduce a Stochastic Gradient Shortcut layer that allows the decoder to utilize the continuous latent vectors directly during training.
    }}
    \label{fig:shape_tokenization}
\end{figure}

 We need a representation of 3D geometry that is expressive enough to capture a wide range of geometric properties faithfully, including smooth surfaces, sharp edges, high-frequency details, and can serve as both input and output tokens to multi-modal autoregressive sequence models. Faced with these needs, we start from an expressive, continuous shape representation such as 3DShape2VecSet~\citep{zhang20233dshape2vecset}, and adapt it into discrete tokens to enable native handling of input and output tokens across multiple modalities in a manner similar to mixed-modal foundation models such as Chameleon~\citep{team2024chameleon}.

As outlined in Figure~\ref{fig:shape_tokenization}, our high-level architecture follows an encoder-decoder design that encodes the input 3D mesh into a latent representation capable of being decoded into an implicit occupancy field. One key distinction is that the continuous latents get discretized through an additional vector quantization process, which poses unique training challenges due to its non-differentiability nature. We propose two techniques, namely stochastic gradient shortcut and self-supervised latent space regularization, to address the VQ training challenges ( Section~\ref{sec:method:vq:stochasticgradient} and~\ref{sec:method:vq:regularization}, respectively).  Another architecture improvement is the use of phase-modulated positional encodings that improve the perceiver-based transformer's ability to disambiguate spatially distinct points in cross-attention layers (Section~\ref{sec:method:vq:positional}). These architecture changes allow us to train a shape tokenizer that faithfully captures a wide range of shapes while yielding discrete tokens that are suitable as a new modality for training token-based mixed-modal foundation models.

\begin{figure}
    \centering
    \begin{subfigure}[b]{0.45\textwidth}
        \centering
        \includegraphics[width=\textwidth]{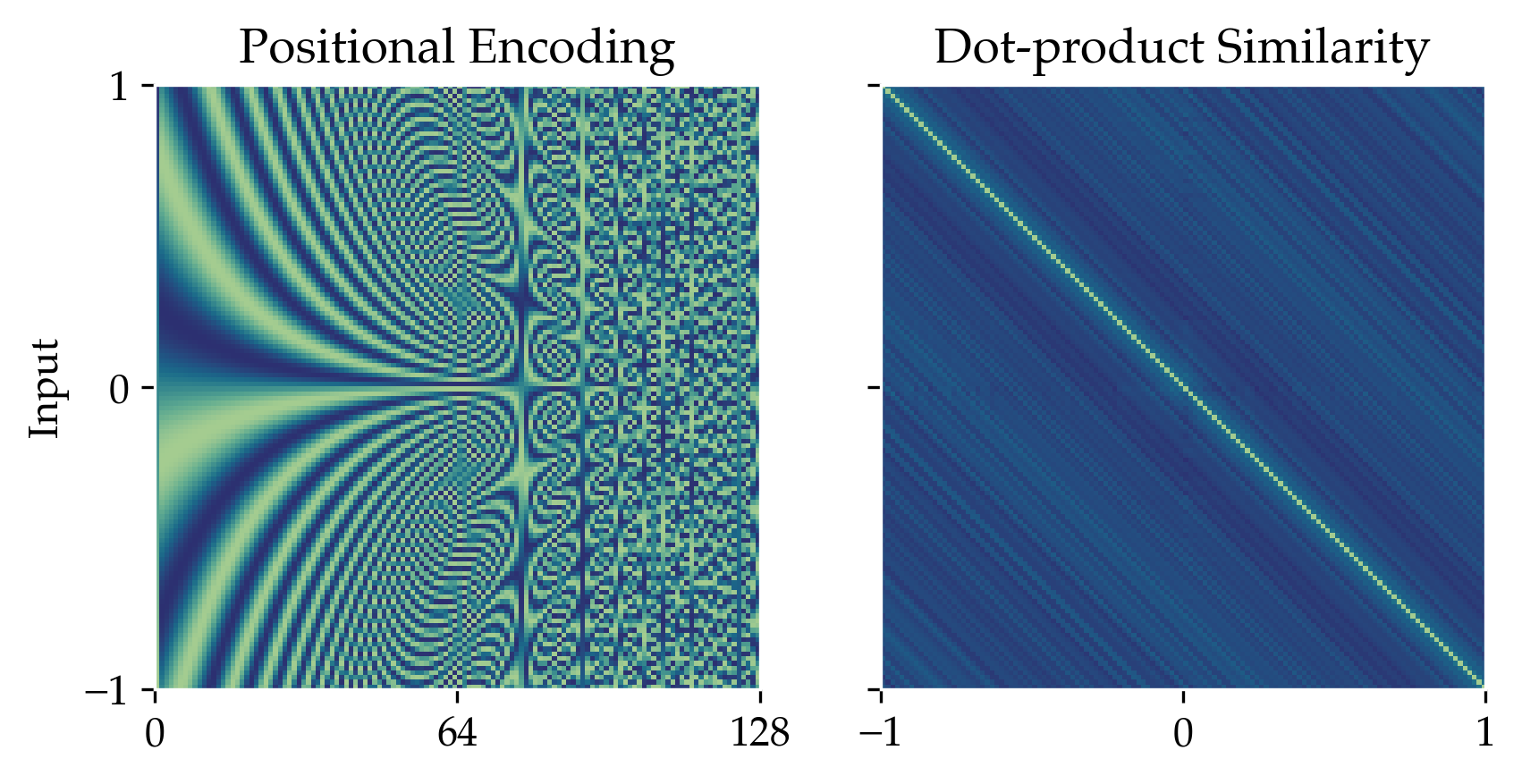}
        \caption{Baseline Positional Encoding $\gamma(\cdot)$}
        \label{fig:pmpe_pe}
    \end{subfigure}
    \hfill
    \begin{subfigure}[b]{0.45\textwidth}
        \centering
        \includegraphics[width=\textwidth]{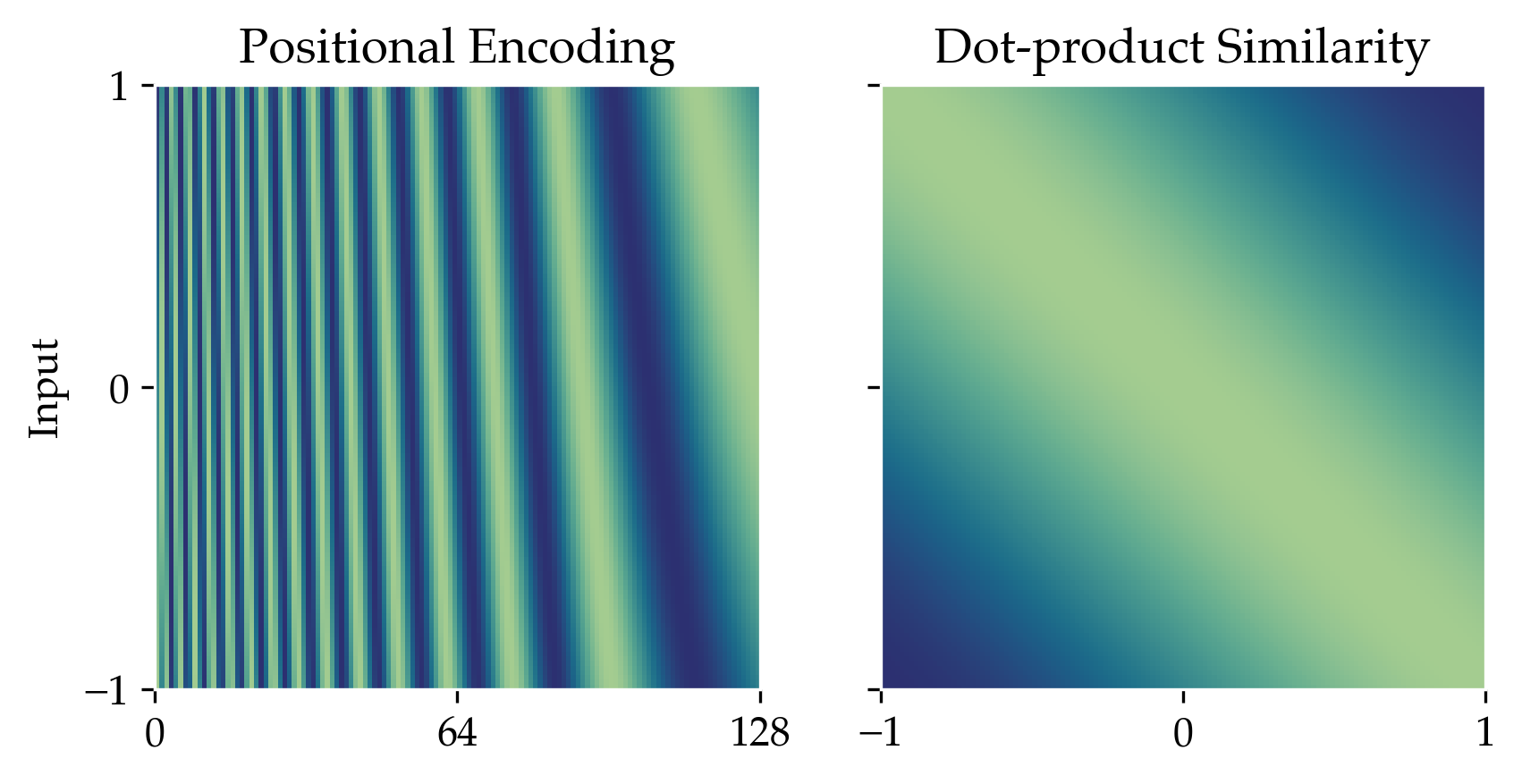}
        \caption{Phase-Modulated Positional Encoding $\gamma'(\cdot)$}
        \label{fig:pmpe_pmpe}
    \end{subfigure}
    \caption{\textbf{Phase-Modulated Positional Encoding.} \small{Comparison of positional encoding methods using 128 base frequencies for 1D inputs in $[-1, 1]$, with corresponding dot-product similarity matrices. \textbf{(a) Traditional positional encoding}  $\gamma(\cdot)$ enhances high-frequency details but exhibits periodic variations across embedding dimensions as inputs vary from $-1$ to $1$. This causes dot-product similarities between encoded vectors to poorly reflect their spatial proximity. \textbf{(b) Our proposed PMPE} $\gamma'(\cdot)$ maintains clear distinctions between spatially distant inputs, as reflected in the dot-product similarity of encoded vectors. Our final approach uses the combined encoding of $\gamma' + \gamma$ that preserves fine-grained details while ensuring similarity patterns align with global spatial proximity.}}
    \label{fig:pmpe}
\end{figure}

\subsection{Phase-Modulated Positional Encoding}
\label{sec:method:vq:positional}

To encode a shape into a compact latent representation, we first sample $N_p$ points from its surface to create a point cloud $\mathbf{P}\in\mathbb{R}^{N_p\times3}$ . Prior work~\citep{zhang20233dshape2vecset,li2024craftsman,zhao2025hunyuan3d,zhao2023michelangelo} embeds $\mathbf{P}$ via a sinusoid positional encoding function $\gamma(\cdot)$~\citep{mildenhall2020nerf} before processing it with a transformer network: 

\begin{equation}
\gamma(p)=\left[\sin{(\omega_1p+\varphi_1)},\ \sin{(\omega_2p+\varphi_2)}, \cdots, \sin{(\omega_Lp+\varphi_L)}\right],
\end{equation}
where $\gamma(p)$ is applied separately to each of the three coordinate channels $p \in [x,y,z]$ in $\mathbf{P}$, and
$\omega_i=2^{\lfloor i/2 \rfloor}\pi,\, \varphi_i=\frac{\pi}{2}(i\, \text{mod}\, 2)$, for $i = 1, \cdots, L$, with $L$ being the number of base frequencies.

The periodicity of $\gamma(\cdot)$ maps points separated by multiples of $2\pi/\omega_i$ in space to identical encodings in the $i^{th}$ channel. The result is that spatially distant points can map to similar embedding vectors (Figure~\ref{fig:pmpe_pe}) that are hard to distinguish after the dot-product in cross-attention layers. The inability for the embedding to disambiguate spatially distant points, and correspondingly, to disambiguate different shape surface features, results in reduced shape reconstruction quality.

To address this problem, we need a new technique to encode points such that it not only captures the multi-scale features as in the traditional position encoding but preserves the distinctiveness of spatially distant points with the dot-product attention. We take inspiration from phase modulation techniques~\citep{haykin2008communication} and propose Phase-Modulated Position Encoding (PMPE). PMPE modulates phase offsets across all the sinusoidal functions,  and uses the embedding function $\gamma_{\text{PM}}$, defined as:

\begin{equation}
\begin{gathered}
\gamma_{\text{PM}}(p) = \gamma(p) + \gamma'(p) , \\
\gamma'(p) =\left[\sin{(\frac{\pi}{2} p+\varphi'_1)},\ \sin{(\frac{\pi}{2} p+\varphi'_2)}, \cdots, \sin{(\frac{\pi}{2} p+\varphi'_L)}\right], \\
\varphi'_i=2\pi\left((\beta L)^{1-\frac{i}{L}}+\frac{i}{L}\right), \quad
i = 1, \cdots, L,
\end{gathered}
\end{equation}
where $\gamma(p)$ is the traditional encoding function, and $\beta$ is a hyperparameter that controls the rate of phase variation across channels. The term $(\beta L)^{(1-\frac{i}{L})}$ here is to vary the base frequency to avoid resonance between $\gamma(p)$ and $\gamma'(p)$.

In contrast to $\gamma(p)$, which uses exponentially increasing frequencies to capture multi-scale features, $\gamma'(p)$ uses the same frequency $\pi/2$ for each channel but varies the phase offset $\varphi'_i$ via a non-linear function of $i$. The non-linear phase modulation ensures the distinctiveness of spatially distant points when mapped to the embedding space, as shown in Figure ~\ref{fig:pmpe_pmpe}. 

Empirically, we find that PMPE yields significantly improved reconstruction fidelity, particularly for complex geometric details. PMPE also produces fewer artifacts such as disconnected components that appear in methods that use only traditional positional encoding.

\begin{figure}
    \centering
    \includegraphics[width=\linewidth]{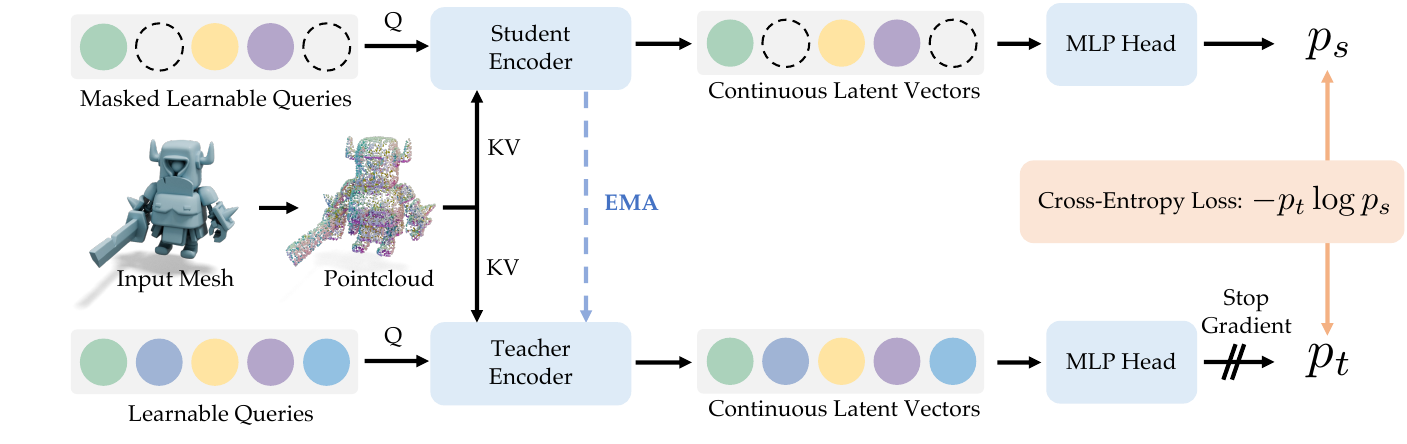}
    \caption{\textbf{Latent Space Regularization with Self-supervised Loss.}
    \small{We regularize our latent space using a self-supervised loss inspired by DINOv2~\citep{oquab2023dinov2}. This loss is computed as the cross entropy between features extracted from the student and teacher encoders, where the teacher model is an Exponential Moving Average (EMA) of the student. Input queries to the student encoder are randomly masked for robustness.
    }}
    \label{fig:self_supervised_loss}
\end{figure}

\subsection{Stochastic linear shortcut for gradient stabilization}
\label{sec:method:vq:stochasticgradient}

After encoding the input shape into a continuous latent vector, we employ a variant of VQ-VAE~\citep{VQVAE} (i.e., Optimal Transport VQ~\citep{zhang2024optvq}) to convert the latents into a sequence of discrete tokens. The quantization layer in VQ-VAE can introduce training instabilities due to ill-defined gradients from the non-differentiable code embedding assignment~\citep{huh2023vqtorch}. Prior works have attempted to use a straight-through estimator~\citep{bengio2013ste} or randomly replace a fraction of the quantized embeddings with their continuous approximations during training~\citep{regularized_vq,sqvae}. Inspired by the prior works, we make a further adjustment by introducing an additional \textbf{linear shortcut layer} that stochastically skips the entire quantization bottleneck. 

With $50\%$ probability, we project the continuous latents from the encoder with a linear layer, and directly feed the projected latents to the decoder, bypassing the quantization layer entirely. Notice that this is different from the direct shortcut approach analyzed in the previous work~\citep{rotationtrick} that is equivalent to using an identity (instead of linear) layer in our formulation. Both~\cite{rotationtrick} and our own experiments have found the direct shortcut approach to not perform well. Our intuition is that the additional linear layer in our formulation allows the shortcut pathway to learn a slightly different set of weights but with well-defined gradients, and it allows the shortcut pathway to act as a teacher network for the quantization pathway. As a result, the quantization pathway is less prone to being stuck at a local minima. Empirically, we observe that incorporating the linear shortcut layer leads to a lower training and validation loss and makes training more stable under a broader range of hyperparameters.

\subsection{Learning geometrically clustered latents with self-supervised loss}
\label{sec:method:vq:regularization}

\begin{figure}
    \centering
    \begin{subfigure}[b]{0.45\textwidth}
        \centering
        \includegraphics[width=\textwidth]{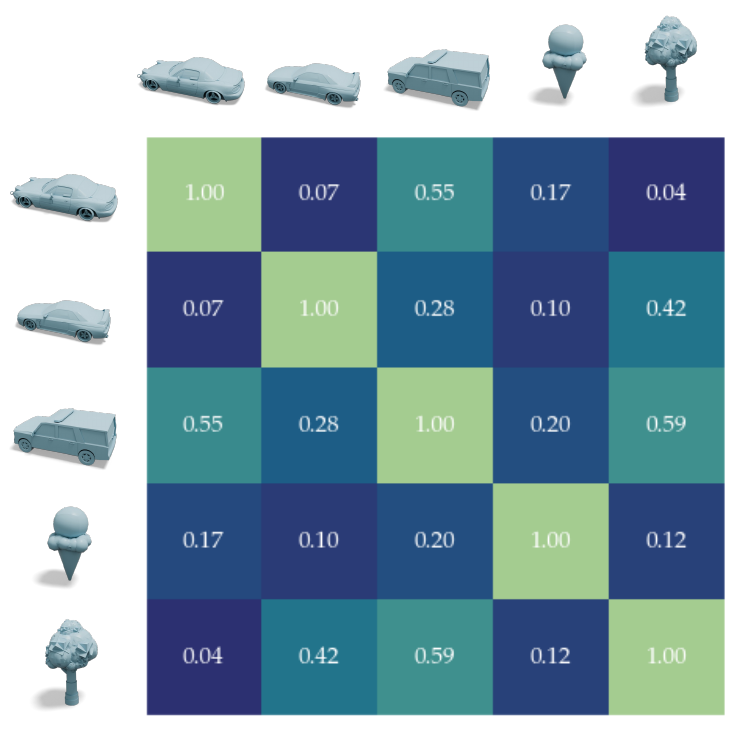}
        \caption{Without self-supervised loss}
        \label{fig:self_supervised_loss_comparison_a}
    \end{subfigure}
    \hfill
    \begin{subfigure}[b]{0.45\textwidth}
        \centering
        \includegraphics[width=\textwidth]{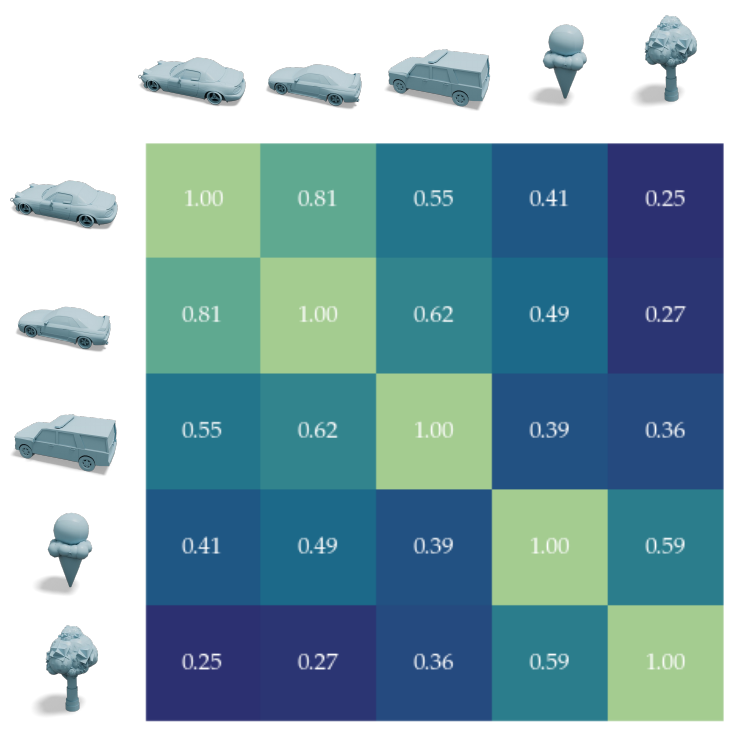}
        \caption{With self-supervised loss}
        \label{fig:self_supervised_loss_comparison_b}
    \end{subfigure}
    \caption{\textbf{Impact of Self-Supervised Loss.} \small{Cosine similarity between the latent vectors of 3D meshes for models trained (a) with and (b) without the proposed self-supervised loss term. \textbf{(a)} Without regularization, latent cosine similarity scores fail to correlate with geometric structure (e.g., the ice-cream in $4^{th}$ row has higher similarity to the car in $1^{st}$ row than the similar car in $2^{nd}$ row). \textbf{(b)} With the loss, latent space similarity aligns with ground-truth geometric relationships.}}
    \label{fig:self_supervised_loss_comparison}
\end{figure}

Following prior work in vision models~\citep{grill2020bootstrap,caron2021emerging,zhou2021ibot,oquab2023dinov2},
we employ a self-supervised loss to regularize the latent space to be geometrically clustered and smooth, where similar shapes produce latent vectors that are close to each other. Figure~\ref{fig:self_supervised_loss} illustrates our encoder's self-supervised learning pipeline. Following the architecture in DINOv2~\citep{oquab2023dinov2}, we maintain an exponential moving average version of the encoder as the teacher model. The student encoder receives randomly masked input queries, while the teacher encoder has access to the full set of queries. Both encoders use an additional MLP head to produce ``prototype scores'', and the self-supervised loss is the cross-entropy between these scores, weighted by a hyperparameter $\lambda_{SSL}$ that balances with the occupancy reconstruction loss. As shown in Figure~\ref{fig:self_supervised_loss_comparison}, the ability to encode geometrically similar shapes into latent vectors with large cosine similarities naturally emerges from the additional self-supervised loss. We anticipate that this property will prove valuable for a wide range of shape processing applications.

\subsection{Results}

\noindent\textbf{Architecture details.}~~For training our shape tokenizer, we use similar transformer layers for both the encoder and the decoder, with each layer having $768$ width  and $12$ heads. The encoder and decoder consist of $13$ and $24$ transformer layers respectively, with a total number of $273$M parameters. We use $512$ latent code tokens and a codebook size of $16,384$ with $32$-dimensional embeddings. We set $\beta=0.125$ for PMPE and $\lambda_{SSL}=0.0005$ for the self-supervised loss. For the VQ layer, we use a recent variant called OptVQ~\citep{zhang2024optvq} that integrates an optimal transport method for index assignment. 

\noindent\textbf{Training data.}~~We train our shape tokenizer on roughly $1.5$ million 3D object assets that utilize a combination of licensed and public available datasets (e.g. Objaverse~\citep{objaverse}), as well as the free data assets that are available in Roblox Creator Store that are opted in for training. We center and rescale all assets to lie within a normalized bounding box with each axis ranging from $[-1, 1]$. For each asset within a training batch, we sample $8,192$ points directly on its surface to facilitate input encoding. To compute the occupancy loss, we sample another $8,192$ points comprising two different sets, one from uniform sampling in the space to capture the overall geometric structure and the other from near-surface to capture high-frequency surface details.

\begin{figure}
    \centering
    \includegraphics[width=\linewidth]{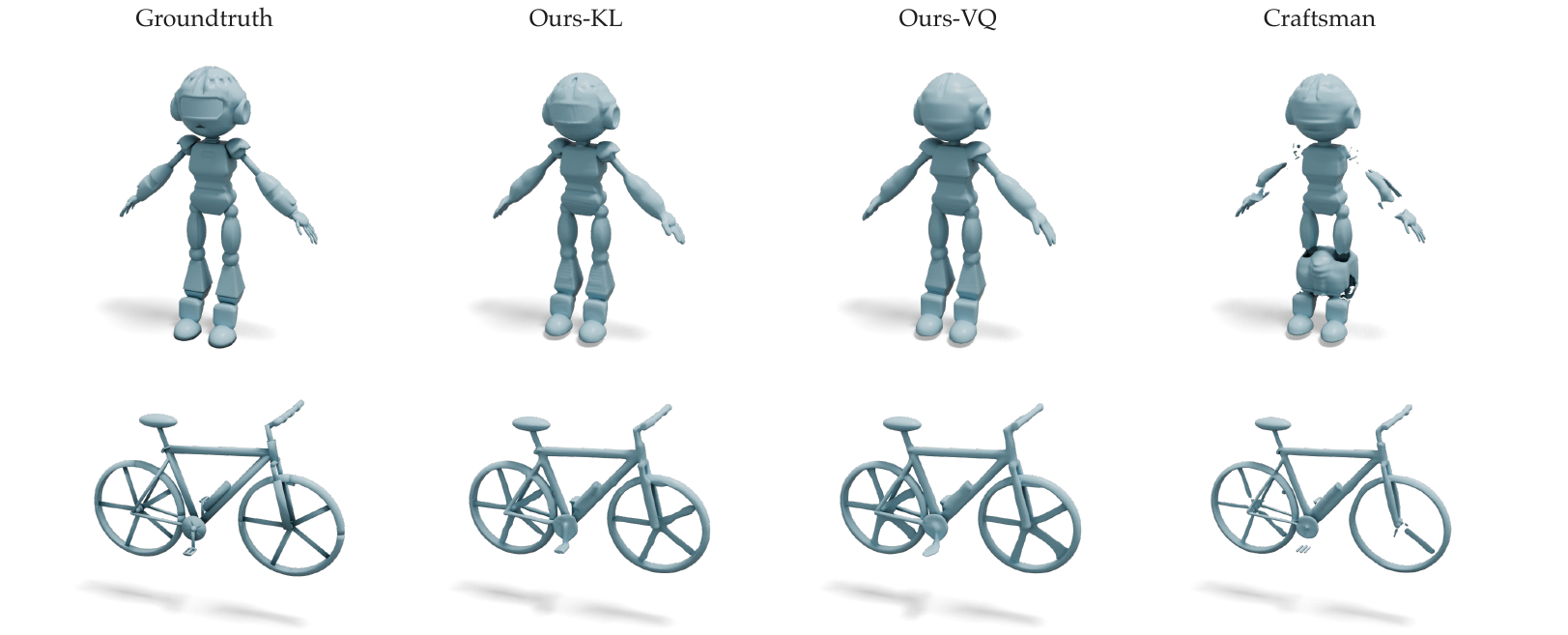}
    \caption{\textbf{Qualitative Analysis of Shape Reconstruction.} Comparison between our method and Craftsman~\citep{li2024craftsman} demonstrates that both of our model variants  achieve superior reconstruction quality, preserving finer geometric details while producing fewer artifacts.
    }
    \label{fig:reconstruction_comparison}
\end{figure}

\begin{table}[t]
\centering
\begin{tabular}{lccc}
\toprule
 & \textbf{CraftsMan}~\citep{li2024craftsman} & \textbf{Ours-VQ} & \textbf{Ours-KL} \\
\midrule
\textbf{S-IoU$\uparrow$} & 68.8\% & 91.7\% & 94.8\% \\
\textbf{V-IoU$\uparrow$} & 83.6\% & 94.5\% & 95.4\% \\
\bottomrule
\end{tabular}
\caption{Comparison of S-IoU and V-IoU results across different models.}
\label{tab:siou-viou-comparison}
\end{table}

\noindent\textbf{Model comparison.}~~We compare our discrete shape tokenizer with a continuous variant trained on the same 1.5~million object dataset, and with CraftsMan~\citep{li2024craftsman} a recent variant of 3DShape2VecSet trained on a 170K object subset of Objaverse.
Our continuous variant is trained using the same architecture as the discrete tokenizer, with the only difference that it skips the VQ process and uses KL-divergence to regularize the latent space~\citep{vae}. We measure the shape reconstruction quality of different methods using surface-IoU (S-IoU) and volumetric IoU (V-IoU) on the Toys4K~\citep{toys4k} dataset. We calculate S-IoU using points near the mesh surface and V-IoU using uniformly sampled points within the bounding volume. None of the three methods was trained on the Toys4K dataset. 

As shown in Table~\ref{tab:siou-viou-comparison} and Figure~\ref{fig:reconstruction_comparison}, both our VQ-VAE (Ours-VQ) and the continuous variant (Ours-KL) outperform CraftsMan in both S-IoU and V-IoU metrics.
Our continuous variant still outperforms its discrete counterpart, highlighting that there remains some loss of geometry fidelity through the vector quantization process. We aim to bridge this gap in future work.

\section{Applications}
Shape tokenization is only one component of the unified foundation model we envision, but we have already found it to be an effective representation of geometric data. We have begun developing a variety of applications using our shape tokens as the core geometric representation. 
These applications include text-to-shape generation, shape-to-text generation and text-to-scene generation. We briefly
explain how each application takes advantage of our shape tokenization scheme and
demonstrate how they might aid users in developing 3D experiences.

\begin{figure}
    \centering
    \includegraphics[width=\linewidth]{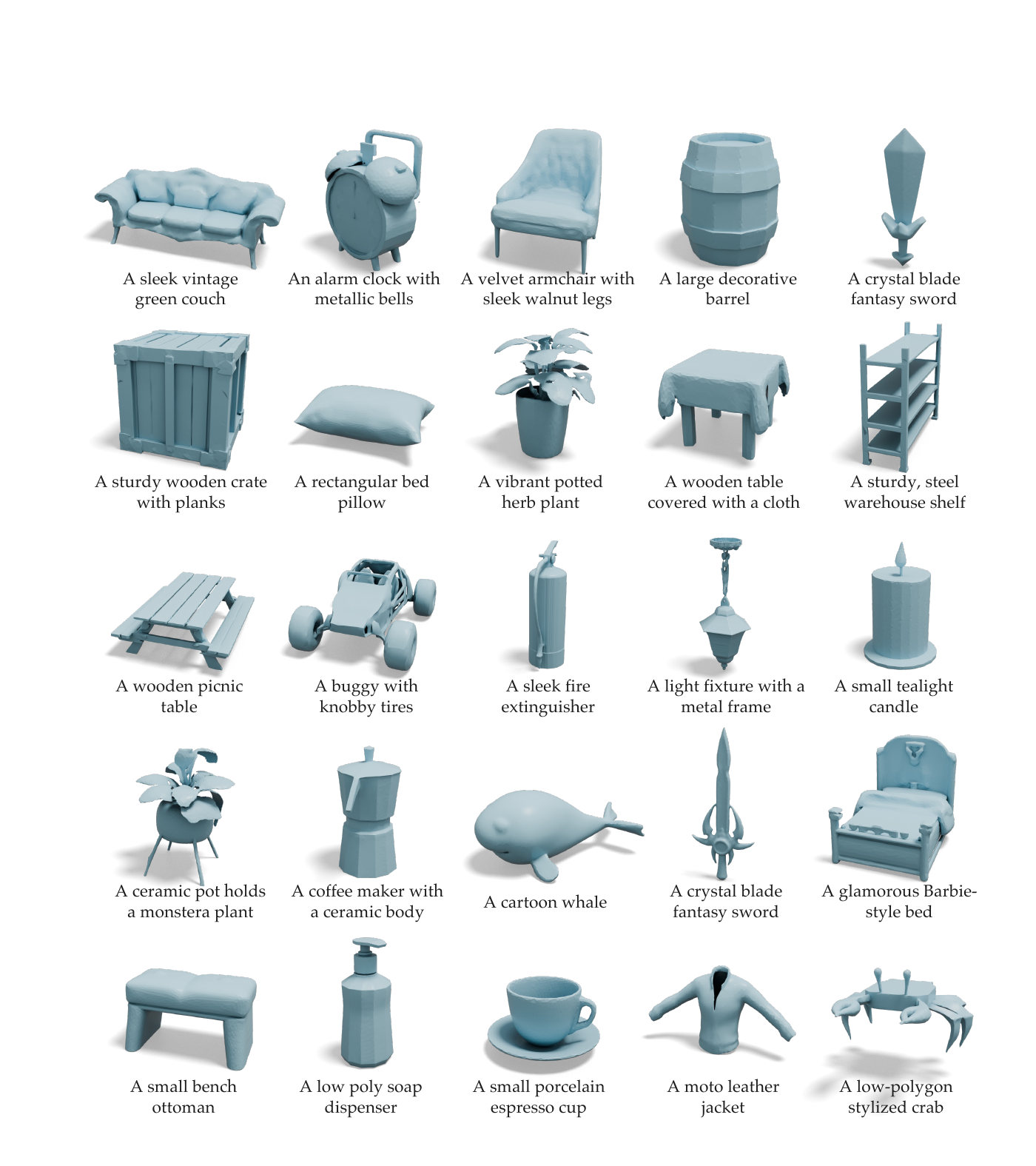}
    \caption{\textbf{Text-to-Shape Generation Result Gallery.} Our model can generate a diverse set of 3D meshes, capturing sharp edges, smooth surfaces and complex structures.}
    \label{fig:generation_results}
\end{figure}

\subsection{Text-to-shape generation}
\label{sec:text-to-shape}

Our text-to-shape application allows users to convert a text description of a shape into a triangle mesh model that can serve as an element of a 3D experience.

\noindent\textbf{Architecture.}~~Our text-to-shape architecture is a decoder-only transformer similar to GPT-2~\citep{radford2019language}. We tokenize the shape into discrete tokens and use the transformer to generate the shape tokens with text conditioning. Specifically, we encode input text prompts using a pre-trained CLIP text encoder~\citep{CLIP}, and leverage a dual-stream attention~\citep{esser2024scaling} to inject the text conditioning into the transformer model to output shape tokens in an autoregressive manner. We also incorporate Classifier-free Guidance~\citep{ho2022classifier} into our GPT training and inference pipeline. During training, we randomly drop the text conditioning and replace it with an empty string 10\% of the time following~\citep{sun2024autoregressivemodelbeatsdiffusion}.

\noindent\textbf{Training data.}~~Training our text-to-shape application requires paired examples of text prompts and their corresponding 3D shapes. For each of the $1.5$ million assets used to train the shape tokenizer, we render multiple views and use GPT-4o~\citep{achiam2023gpt} to provide captions of various lengths.

\noindent\textbf{Mesh extraction.}~~We use marching cubes~\citep{lorensen1998marching} to extract the iso-surface from the occupancy field, along with an in-house mesh decimation algorithm built upon the quadric errors~\citep{garland1997surface} for simplifying the mesh into desired number of faces. Finally, we apply a post-processing step that removes small disconnected components (i.e. floater artifacts) from the output mesh.

\noindent\textbf{Results.}~~We show a representative set of text-to-shape generations in Figure~\ref{fig:generation_results}. Our text-to-shape model is able to generate a diverse set of 3D meshes, capturing sharp edges, smooth surfaces and complex structures. These meshes, as produced with our discrete shape tokens are approaching the visual quality to those produced using recent methods based on rectified flow transformers with continuous latents such as Trellis~\citep{xiang2024trellis}, Hunyuan3D-2~\citep{zhao2025hunyuan3d} and TripoSG~\citep{li2025triposg}. While a benefit of our approach is that it can easily support multi-modal tokens, a full understanding of the relative strengths of the different approaches is an area of future work.

\begin{figure}[t]
    \centering
    \includegraphics[width=\linewidth]{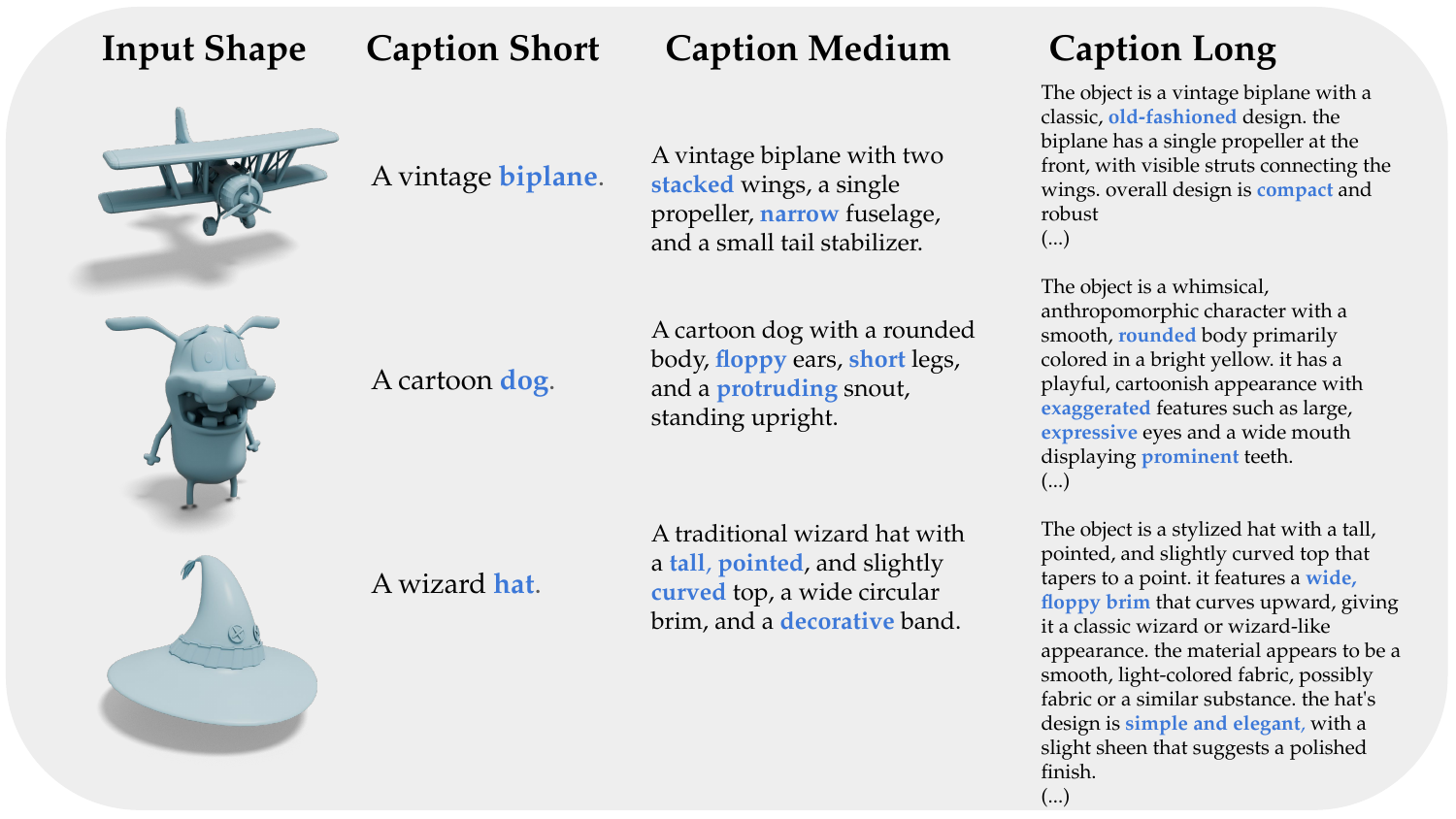}
    \caption{\textbf{Shape-to-text examples.} \small{Shape-to-text captioning of example shapes from Toys4K dataset using short, medium and long captions. Words highlighted in blue indicate notable differences as captions increase in length. Short captions usually capture the shape category. More details about the shape's geometry and style are added as caption length increases.
    }}
    \label{fig:shape2text}
    \vspace{-2mm}
\end{figure}

\begin{figure}[t]
    \centering
    \includegraphics[width=\linewidth]{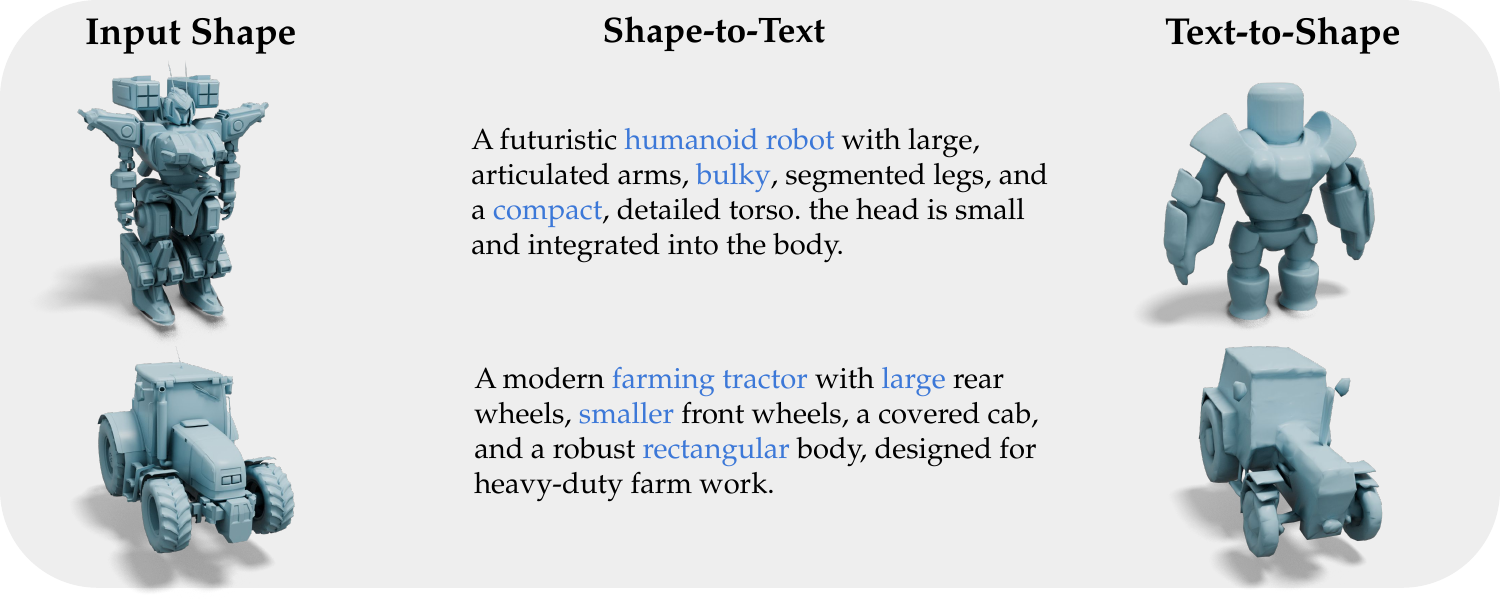}
     \caption{\textbf{Shape cycle consistency.} \small{Our shape-to-text and text-to-shape models demonstrate the cycle consistency. Given a shape, we caption it using shape-to-text model, then then regenerate the 3D shape using the text-to-shape model. We highlight some keywords in blue for the shape category and style. The process preserves the overall geometry and key characteristics of the original shape, although there can be some loss in fine-scale details. 
    }}

    \label{fig:shape_consistency}
    \vspace{-2mm}
\end{figure}

\subsection{Shape-to-text generation}
\label{sec:shape2text}

Our shape-to-text application produces a descriptive natural language caption for an input 3D shape. These text captions are designed to capture 3D shape information in the sense that they can be fed back to our text-to-shape application to produce corresponding 3D models. 
And, as we will show in Section\,\ref{sec:scenegen}, these text captions allow us to build
AI tools that can seamlessly collaborate with existing LLMs to leverage their common sense knowledge and reasoning capabilities.

Our architecture is inspired by recent work on vision-language models such as LLaVA~\citep{liu2023visualinstructiontuning}. In order to perceive visual signals in vision-language models, a pre-trained openset visual encoder, such as CLIP\,\citep{CLIP}, is commonly used to encode the visual input into the input space of the language decoder. In our case, we encode the 3D shape input with our shape tokenizer and inject the shape tokens into a pre-trained decoder-only transformer~\citep{vaswani2017attention} with a single two-layer MLP
latent projection layer. For the pre-trained LLM backbone, we chose the language model from InternVL 2.5-2B~\citep{chen2025expandingperformanceboundariesopensource}, a multimodal large language model pre-trained on large-scale image-text data.
We choose this backbone due to its outstanding performance on a wide range of multimodal benchmarks. Our architecture design follows a standard multimodal learning practice and it demonstrates that our shape tokenizer can be adapted and applied in standard multimodal setups.

\textbf{Training}. 
Since the shape tokenizer is trained on the shape dataset only, the shape latent space can be different from that of the language transformer. Following the training recipe of LLaVA\,\citep{liu2023visualinstructiontuning}, we adopt a two-stage training pipeline.
In the first stage, only the latent projection layer is trained to align the shape features with the text features of the pre-trained transformer. In the second stage, we jointly finetune the latent projection layer
and the weights of the language transformer to better align the shape features with the text features. We keep the shape tokenizer frozen during the entire training process. Similar to LLaVA, we train the model using the next-token prediction loss and only apply the loss to the text output sequence.

\textbf{Implementation details.} We use the same dataset we developed for text-to-shape generation (Section\,\ref{sec:text-to-shape}). In the dataset, we have captions of various length. In order to control the output caption length in the shape-to-text generation, we append a short text instruction after the input shape tokens to indicate the desired output caption length. Specifically, we use \textit{``caption short:``}, \textit{``caption medium:``}, and \textit{``caption long:``} for caption lengths of less than 25 tokens, less than 75 tokens, and greater than 75 tokens respectively.

\textbf{Results.} Figure~\ref{fig:shape2text} shows captions of various lengths generated by our shape-to-text model. When producing short captions, the model only gives category information about the input shape (e.g., a cartoon dog). As the caption length increases, the model describes different parts of the object (e.g., rounded body, floppy ears, etc.) and styles (e.g., playful, cartoonish appearance, large, expressive eyes, etc.).

\textbf{Shape cycle consistency.} Our goal is to capture a notion of 3D shape in the natural language captions produced by our shape-to-text model. 
Figure~\ref{fig:shape_consistency} shows examples where we start from shape input, generate corresponding text captions using our shape-to-text model, and then regenerate the shape by treating the captions as inputs to our text-to-shape model. The results show that the captions produced by our shape-to-text model capture enough 3D information for our text-to-shape model to reproduce the overall 3D shape, albeit with some loss of high-frequency geometric details. We hypothesize that shape cycle-consistency emerges from the fact that the same shape tokens are used by the shape-to-text and text-to-shape models.

\subsection{Text-to-Scene generation}
\label{sec:scenegen}
Our text-to-scene application combines the text-to-shape and shape-to-text models from the prior sections with an off-the-shelf LLM to provide users the ability to iteratively construct full scenes using only text descriptions. From simple prompts like ``Make a campsite'' and also corrective instructions like ``Add another tent'', our scene generation application creates a collection of objects and lays them out (sets position, orientation and scale) in a scene configuration. The system can also make suggestions to the user about what additional elements should be added to the scene (e.g., consider adding neon lights to a 60's-style diner scene).

\definecolor{bggray}{rgb}{0.95,0.95,0.95}

\begin{figure}
    \centering
    \includegraphics[width=\linewidth]{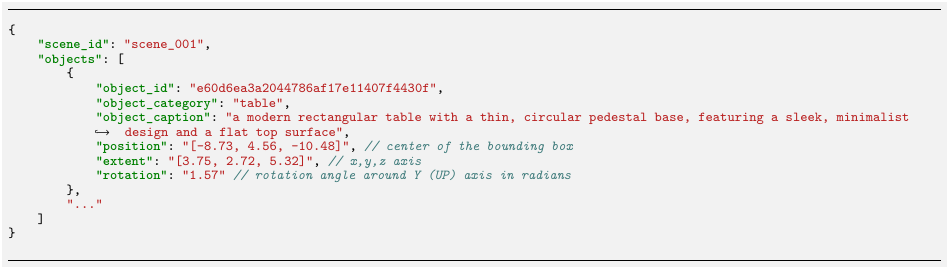}

\caption{\textbf{Scene graph with text-based object descriptions.} \small{For each scene object, the scene description includes the type of object (\texttt{object\_category}) and a text description of its shape (\texttt{object\_caption}). These descriptions facilitate LLM understanding of the scene. To support the simple case of orienting objects on a ground plane, our current implementation only supports Y axis rotations.}}
\label{fig:scene-schema}
\vspace{-2mm}
\end{figure}

{\bf Scene graph representation.} Our shape-to-text model allows us to convert entire scene descriptions (the individual object shape descriptions and description of their placement in a scene) into a single text-based scene-graph that can be consumed and processed by LLMs. Our scene graph format is JSON-based, and represents a scene as a flat list of 3D objects (Figure~\ref{fig:scene-schema}). Each object includes a text description of the object's shape ({\tt object\_caption}) and the type of object it is ({\tt object\_category}).
These fields provide a sufficiently detailed description of the 3D object for an LLM to be able to reason about some of its spatial and semantic properties.
Objects also have scene layout information giving the position, orientation, and bounding box of the object in the scene.
Given a 3D scene (such as via a GLTF file), we convert the scene to our textual scene graph by extracting all references to 3D models and replacing them with their bounding box and the appropriate text descriptions produced by our shape-to-text model (Section~\ref{sec:shape2text}). Conversely, given a scene graph containing only a textual description of objects, we can convert all textual object descriptions into 3D geometry using text-to-shape (Section~\ref{sec:text-to-shape}).

\begin{figure}
    \setlength{\abovecaptionskip}{-10pt}
    \centering
    \vspace{-2mm}
    \includegraphics[width=0.95\linewidth]{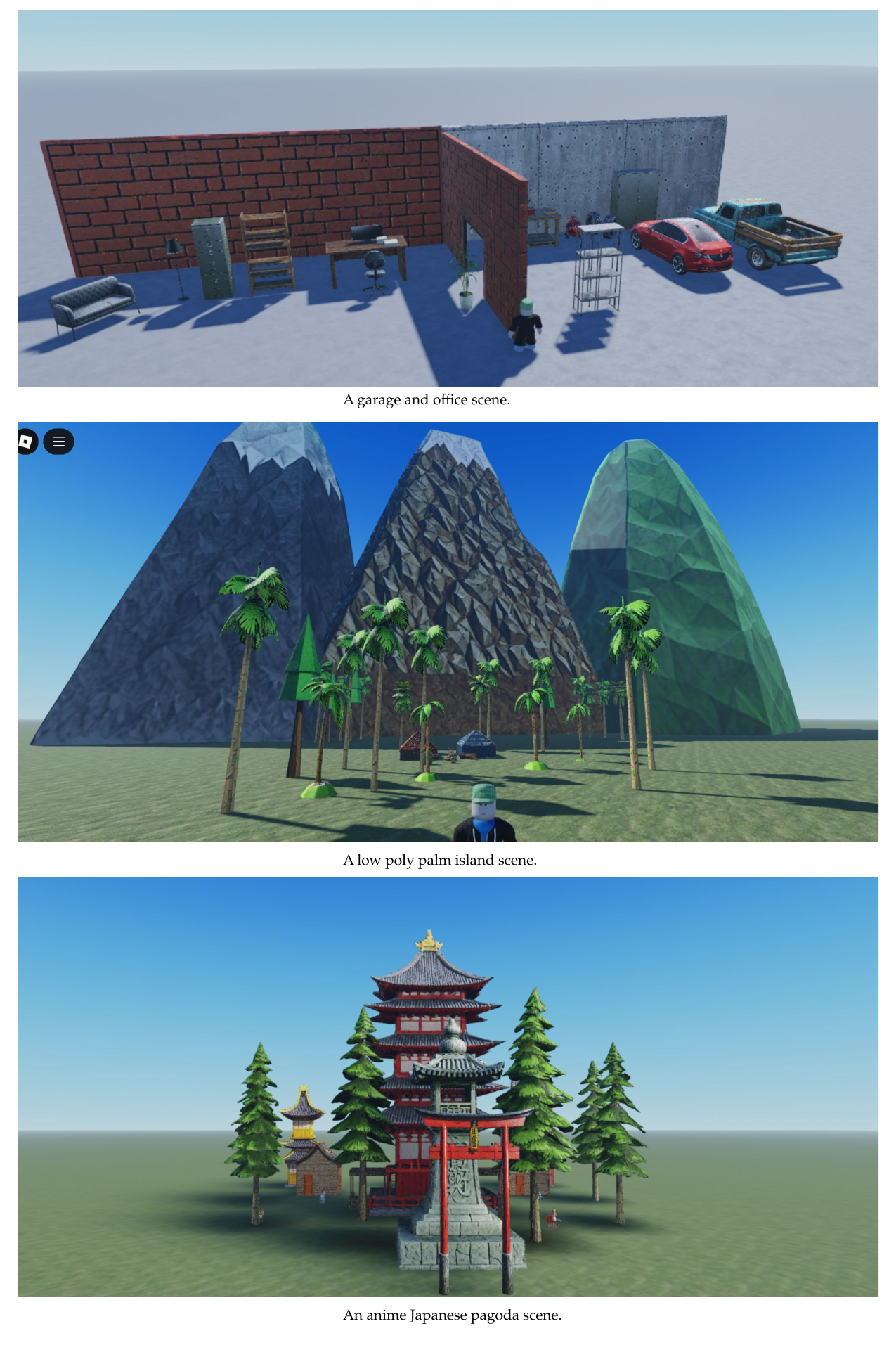}
    \caption{\textbf{Scene generation examples.} \small{The layout, the rotation, position and scale of each object are produced by the LLM system, while the geometry and texture are generated by our text-to-shape and an in-house text-to-texture model built upon FlashTex~\citep{deng2024flashtex}. 
    }}
    \label{fig:scene_example}
    \vspace{-2mm}
\end{figure}

\begin{table}[t]
    \centering
    \begin{minipage}{1.0\textwidth}
    
    \renewcommand{\arraystretch}{1.0} %
    \setlength{\tabcolsep}{4pt} %
    {\footnotesize
    \begin{tabularx}{\textwidth}{lX}  %
        \toprule
        
        \multicolumn{2}{l}{\small\textbf{3D scene suggestion assistant: }} \\  
        \midrule
        \multicolumn{2}{c}{\includegraphics[width=8cm]{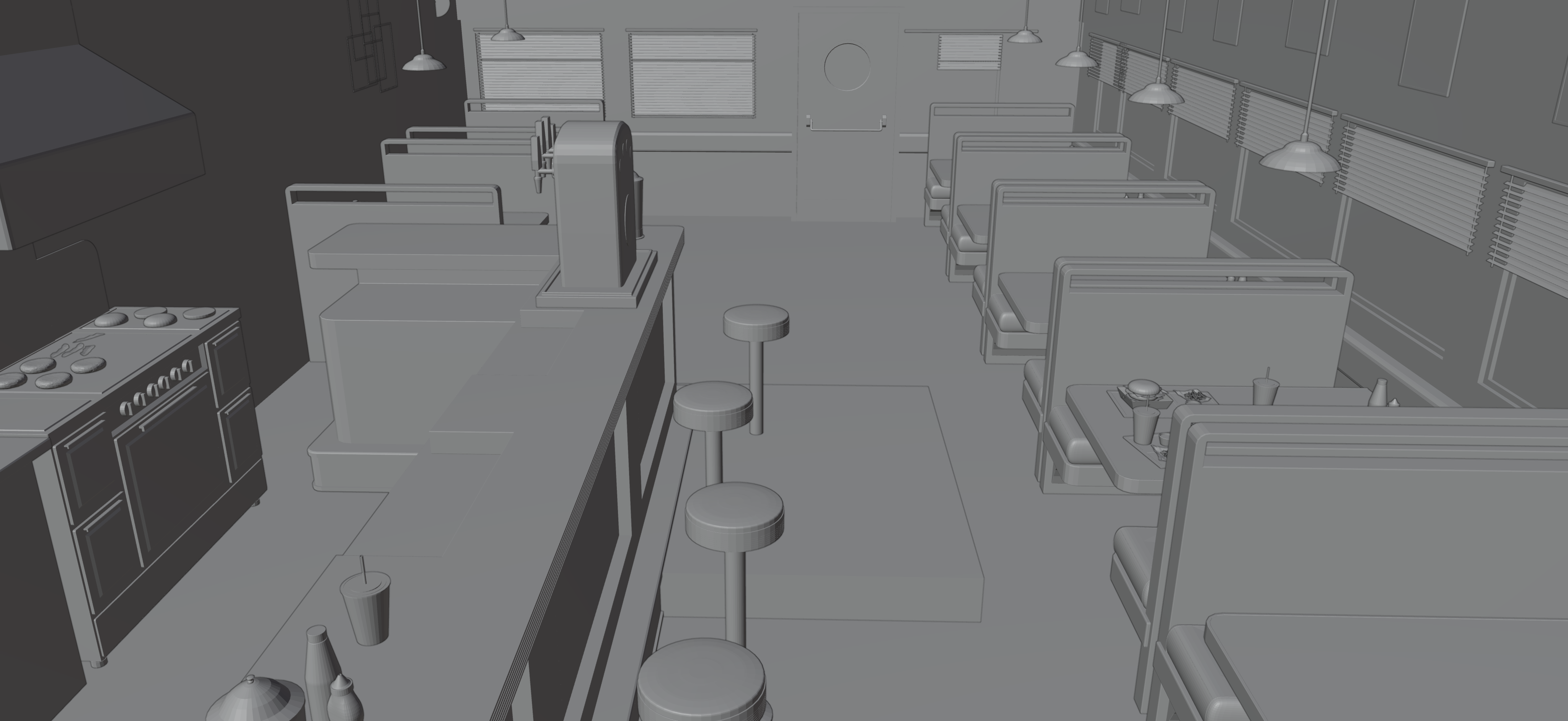}} \\ %
        &\multicolumn{1}{X}{\tiny{(We remove all the scene related meta data, i.e. scene name, object category name from the artists, and only use the shape caption generated by our shape-to-text model to test the 3D scene Q\&A)}}\\
        \midrule
        \textbf{User} & Please summarize the scene. \\
        \textbf{Assistant} & The scene features a \textbf{modern dining} setting with \textbf{tables}, \textbf{chairs}, and various objects like \textbf{plates, bottles, coins}, and decorative items. Pendant lights hang above, and a sleek, minimalist aesthetic dominates the space. There are also some irregularly shaped objects, possibly \textbf{food items}, enhancing the dining atmosphere. \\
        
        \midrule
        \textbf{User} & I’d like to add condiments on top of the counter, where should I place them? \\
        \textbf{Assistant} & The scene includes modern \textbf{rectangular} tables and a \textbf{counter-like} surface (likely the trapezoidal prisms or modern tables). Consider these positions: 1.\textbf{Near the center of the counter} – Ensures easy access for diners. 2.\textbf{Close to plates or food items} – Keeps condiments within reach of those eating. 3.\textbf{Evenly spaced along the counter} – Prevents crowding in one area.
 \\
        \midrule
        \textbf{User} & I want to replace the stool near the counter, what other type of seat do you recommend? \\
        \textbf{Assistant} & For a modern dining scene, here are some great alternatives to replace the stool near the counter: 1.\textbf{High-Back Bar Chairs} – Sleek and cushioned for a premium feel. 2.\textbf{Bench Seating} – A continuous, modern look that fits multiple players. 3.\textbf{Swivel Chairs} – Adds a dynamic, interactive element.
 \\
        \midrule
        \textbf{User} & What audio, background music would you recommend in this scene? \\
        \textbf{Assistant} & For your modern dining scene, I recommend:
        1. \textbf{Lo-fi jazz} with soft piano \& muffled chatter for a chic, relaxed vibe. 2. \textbf{Light orchestral music} with faint dish clinks for an elegant touch. 3. \textbf{Acoustic guitar \& café ambiance} for a cozy feel.
\\
        \bottomrule
    \end{tabularx}
}
    \end{minipage}
    \caption{\textbf{Scene analysis and suggestion example.} \small{Given this diner scene, our scene suggestion assistant is able to summarize the scene, make suggestions for scene enhancements such as where to placement of condiments, what style of seating would be appropriate, and make background audio recommendations.}}
    \label{fig:scene_und}
    
\end{table}

{\bf Making a scene.} Given a user-specified scene description, our text-to-scene application uses an LLM to generate scene descriptions in the format described above. Modern LLMs exhibit strong capabilities for reasoning about what objects are likely to be present given the input description, but they can struggle with more precise spatial reasoning needed to construct plausible scene layouts. To assist the LLM we generate a database of prompt and JSON scene graph pairs (created by processing scenes using our shape-to-text model described above) and provide exemplars from this database to the LLM via in-context learning\,\citep{dong2022survey}.  Once the LLM has generated a JSON scene graph, we generate scene geometry for each object using our text-to-shape model.
Figure~\ref{fig:scene_example} shows renderings of scenes generated using our text-to-scene generator. The generated scenes are diverse and the layout placement is physically-plausible (i.e., the sofa, chair and the office desk are placed in the correct location and the cars in the garage are facing in the consistent direction). The objects in the scene have consistent styles in both the geometry and texture (e.g., the pagoda and the traditional Japanese front gate match in the scale and style). The scenes depicted in the figure were generated by the LLM from a starting scene description prompt, and then iteratively refined by the user through multi-turn conversations with our system.  For example the user can issue new text instructions to the LLM, (``Add a tree to the scene'') resulting in a new scene description, or use a 3D scene editor tool to manually adjust the position and orientation of objects after the initial placement, which can be used to occasionally correct the LLM placement errors.

{\bf Scene analysis and suggestions.} When creating a 3D scene, it can be challenging to consider what objects populate it with. We envision a human user working in tandem with our scene creation application to help them to quickly consider options for how to improve the scene. For example, Table~\ref{fig:scene_und} depicts a 3D scene of a 60's-style diner. The text description of a scene provides sufficient information for the LLM to comprehend how the scene might be used, and to offer suggestions on how to enhance it. For example, the system is able to suggest relevant places to put condiments on the counter top, recommend styles of seating that might be added, and even offer music suggestions.

\section{Conclusion}

We have demonstrated the first step towards our goal of developing a foundational model that is capable of 3D generation and reasoning. Specifically, we have shown how our shape tokenization scheme enables a variety of generative 3D applications. Looking ahead, there are multiple challenges that must be addressed to achieve the complete vision of a unified 3D foundational model capable of assisting human developers with building all aspects of a Roblox experience. Here we outline some of the future steps towards this vision.

{\bf Mixed generation of meshes and parts.} Our text-to-shape application generates a 3D geometric mesh from a text prompt. But the Roblox platform also supports modeling 3D shapes via constructive solid geometry (CSG) operations on basic parts (e.g. cylinder, cube, sphere, etc.). These parts-based objects provide a compact representation that can be rendered efficiently on a variety of edge devices. They also provide a distinct ``blocky'' visual style that is the hallmark of many Roblox experiences.
We are exploring how to incorporate our shape tokens into an autoregressive GPT model that can generate parts-only CSG shapes or shapes that combine CSG and meshes.

{\bf Character avatar generation.} Avatars are the primary representation of players in Roblox experiences. For many players, their first creation experience involves designing or customizing their avatar. Avatars move and animate as players interact with a scene, so a AI-based avatar generation must produce riggable geometry: detailed head meshes, separate meshes for animatable features such as eyes and mouth, a body mesh, and clothing meshes that are layered on the body. 

{\bf 4D behavior generation.} Objects in a Roblox experience can be rigged for animation and/or programmatically scripted to behave in different ways depending on how a player interacts with them. 
For example, a 3D car model might be rigged so that turning the steering wheel rotates the wheels. A door might be scripted to slide open when a player is near it. We refer to such rigged and/or scripted motions as \textit{4D behavior}, and our goal is to build AI-based 4D behavior generators. 

Our vision of a unified foundational model for 3D intelligence aims to enable users of all backgrounds to easily create complete 3D experiences. Generative AI tools that can assist users at each step of the creation process requires such a foundational model that can reason about 3D shape, object behaviors, player interactions and game logic. Our work so far has taken a step towards the first of these reasoning goals. With the release of our open source model\footnote{ \url{https://github.com/Roblox/cube}} we hope to engage others in the research community to address these goals with us.

\clearpage

\section{Update Logs (July, 2025)}
\label{sec:v2_updates}

Since the initial release in March 2025, we have made some key updates to our open-source text-to-shape model, which we describe in more detail below.

\subsection{Data}
We introduce an additional synthetic data source into the training set. The synthetic data pipeline takes advantage of the generalization capabilities of state-of-the-art text-to-image and image-to-shape models to generate paired $(text, shape)$ data suitable for our training. For generating a diverse set of prompts to be fed into the pipeline, we start with a list of seed prompts that come from two sources: 1) list of objects found within LLM-generated fictional game concepts; 2) real-life objects (e.g., dishwasher, laptop, etc.). Then the seed prompts are fed into an LLM to create multiple, descriptive variations. In total, we collected about $3$ million high-quality synthetic assets for training.

With the addition of synthetic data, we observed substantial improvement in prompt adherence for our text-to-shape model, including for challenging compositional prompts. Please refer to our Github page (\url{https://github.com/Roblox/cube}) for a gallery of updated results.

\subsection{VQ-VAE Shape Tokenization Improvements}
\label{sec:vq_improve}

\begin{figure}
    \centering
    \includegraphics[width=\linewidth]{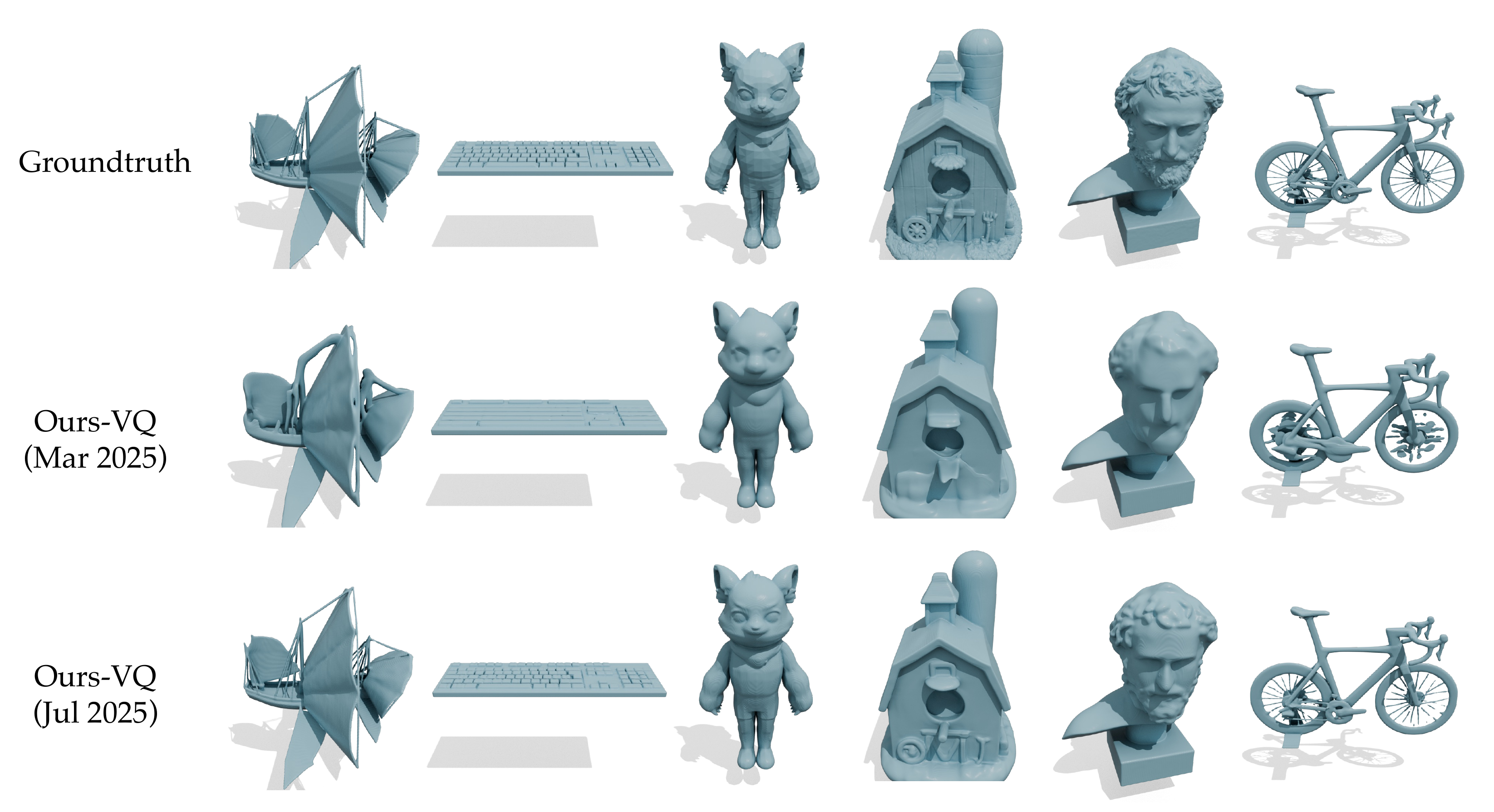}
    \caption{\small{VQ-VAE reconstruction comparison between our previous model (Mar 2025) and our updated model (Jul 2025) with the improvements introduced in Section~\ref{sec:vq_improve}. 
    }}
    \label{fig:vq_improvement}
\end{figure}

To enhance the quality of our VQ-VAE shape tokenization, several key refinements were made to the architecture and training procedure (see sample comparisons with our previous model in Figure~\ref{fig:vq_improvement}). The VQ-VAE latent length was increased from $512$ to $1024$. This expanded VQ-VAE is optimized using a two-stage training protocol along with other regularization and input data improvements detailed below.

\subsubsection{Two-stage training protocol}
Our training process is divided into an initial pre-training stage with occupancy supervision, followed by a fine-tuning stage with Truncated Signed Distance Function (TSDF) supervision. 

\textbf{Stage 1: Occupancy-based pre-training}. The new VQ-VAE model is first trained using supervision from a binary occupancy field, similar to our March'25 model. To accelerate convergence, the $1024$-d VQ-VAE model is initialized from the weights of our initial $512$-d model. All layers are transferred directly, while the new, larger query layer is initialized with the same mean and standard deviation as the $512$-d model.

\textbf{Stage 2: TSDF-based fine-tuning.} Once the occupancy-trained model has converged, it is further fine-tuned using a continuous TSDF. Unlike binary occupancy values, the TSDF provides richer, continuous gradient signals about the surface, enabling the model to capture finer-grained details.

\subsubsection{Regularization and input data}
In addition to the two-stage protocol, we made some more improvements to the training procedure:
\begin{itemize}
    \item \textbf{Eikonal loss}: To enforce a valid SDF, Eikonal loss~\citep{gropp2020implicit} is applied as a regularizer, encouraging smoothness and continuity in the predicted SDF.
    \item \textbf{REPA regularization}: Inspired by the Representation Alignment (REPA) work~\citep{yu2024representation} from the image domain that regularizes learned features with a given semantic representation (e.g. DINOv2~\citep{oquab2023dinov2}), we introduce an additional regularization term in the latent space. This loss minimizes the distance between the VQ quantized latents and the latents just before the final decoder layer of our VQ-VAE. We found that the REPA regularization improves the smoothness of the quantized latents, which is beneficial for the GPT training of the downstream shape generation.
    \item \textbf{Input point cloud density}: To provide the encoder with a more detailed input signal, we increase the number of points sampled from the mesh surface from $8,192$ to $32,768$. This modification enables the shape tokens to capture more surface details without increasing the computation overhead of the subsequent generative model.
\end{itemize}

\subsection{3D Bounding Box Conditioning}

To enable more explicit control of the spatial extent of the generated shapes, we introduce a 3D bounding box as additional input conditioning into our text-to-shape model. The bounding box is represented as a 3D vector containing the unit normalized dimensions of the target shape's axis-aligned bounding box. This vector is fed directly into an MLP, and the output embedding is appended as an additional token along with the text tokens to condition the shape generation.

One challenge that emerged during training is that since each shape is paired with a single deterministic bounding box but can be associated with multiple diverse text prompts, the model learned to prioritize the bounding-box condition, often disregarding the text prompt during inference. To address this, we apply a random perturbation to each dimension of the bounding box during training. This simple data augmentation technique prevents the model from overfitting to the exact dimensions of the bounding box and forces it to balance the influence of both spatial and textual conditional inputs. As a result, the final model remains responsive to the text prompt while adhering to the specified spatial constraints. Please refer to our Github page (\url{https://github.com/Roblox/cube}) for illustrations of the 3D bounding box conditioning. 

\subsection{Accelerated shape extraction}
To accelerate the shape extraction, we make use of a hierarchical volume decoding technique that reduces the computational complexity of sampling the Signed Distance Function (SDF). Standard iso-surface extraction methods, such as Marching Cubes, necessitate evaluating an SDF, $f(\mathbf{x}): \mathbb{R}^3 \to \mathbb{R}$, on a dense volumetric grid of resolution $N^3$, incurring a computational cost of $O(N^3)$. However, the target surface, defined by the zero-level set $S = \{ \mathbf{x} \in \mathbb{R}^3 \mid f(\mathbf{x}) = 0 \}$, implies that its representation within the volume is sparse. Our approach exploits this sparsity by first identifying occupied voxels on a coarse grid of resolution $N_c^3$, where $N_c \ll N$. A voxel is classified as occupied if the SDF values at its vertices exhibit a sign change, indicating the presence of the iso-surface. Voxels determined to be entirely inside ($f(\mathbf{x}) < 0$) or outside ($f(\mathbf{x}) > 0$) the surface are pruned. Subsequently, only the subset of occupied coarse voxels is subdivided to the target resolution $N$ for dense SDF evaluation. This coarse-to-fine strategy constrains the expensive queries to a narrow band surrounding the surface, reducing the complexity from $O(N^3)$ towards $O(N^2)$ and thereby substantially decreasing the number of required cross-attention queries to accelerate inference time.

\clearpage

\bibliography{references}

\newpage
\appendix
\section*{Appendix}

\section{Contributions and Acknowledgments}

\begin{multicols*}{2}
\noindent
\textbf{Team Leads} \\
Kiran Bhat (Overall) \\
Nishchaie Khanna (Infrastructure) \\
Karun Channa (Product) \\
Tinghui Zhou (Modeling) \\

\noindent
\textbf{Contributors} \\
Alexander Weiss \\
Christian Reiser \\
Ravi Kiran Chirravuri \\
Ravali Kandur \\
Alejandro Pelaez \\
Akash Garg \\
Michael Palleschi \\
Jessica Wang \\
Skylar Litz \\
Leon Liu \\
Anying Li \\
David Harmon \\
Derek Liu \\
Liangjun Feng \\
Denis Goupil \\
Lukas Kuczynski \\
Jihyun Yoon \\
Naveen Marri \\
Peiye Zhuang \\
Yinan Zhang \\
Brian Yin \\
Haomiao Jiang \\
Marcel van Workum \\
Thomas Lane \\
Bryce Erickson \\
Salil Pathare \\
Kyle Price \\
Steve Han \\
Yiqing Wang \\

\columnbreak

\noindent
\textbf{Core Contributors} \\
Yiheng Zhu \\
Xiaoxia Sun \\
Charles Shang \\
Anirudh Sudarshan \\
Maurice Chu \\
Daiqing Li \\
Kangle Deng \\
Jean-Philippe Fauconnier \\
Tijmen Verhulsdonck \\

\noindent
\textbf{Research Advisors} \\
Maneesh Agrawala \\
Kayvon Fatahalian \\

\noindent
\textbf{Executive Advisors} \\
Anupam Singh \\
David Baszucki \\

\end{multicols*}

\end{document}